# Complete Scanning Application
# Using OpenCv

Ayushe Gangal, Peeyush Kumar and Sunita Kumari
ayushe17@gmail.com, peeyushstark@gmail.com, sunitakumari@gbpec.edu.in
*Dept. of CSE, GB Pant Govt. Engineering College*
*New Delhi-110020, India*

**Abstract**. In the following paper, we have combined the various basic functionalities provided by the NumPy library and OpenCv library, which is an open source for Computer Vision applications, like conversion of colored images to grayscale, calculating threshold, finding contours and using those contour points to take perspective transform of the image inputted by the user, using Python version 3.7. Additional features include cropping, rotating and saving as well. All these functions and features, when implemented step by step, results in a complete scanning application. The applied procedure involves the following steps: [1] Finding contours [2] applying Perspective transform and brightening the image [3] Adaptive Thresholding and applying filters for noise cancellation [4] Rotation features and perspective transform for a special cropping algorithm. The described technique is implemented on various samples.

1. **Introduction**

There are a number of desirable reasons to migrate from paper-based document system to paper-less document systems. A major reason being that, electronic records are easier to transmit and share remotely. They can also be easily stored and found. They reduce the cost of document storing and transferring, and can easily be backed up. Thus the man-kind, being the most intelligent species, that walked our planet, invented the concept of scanners. The first ever scanner was invented in 1957, by a team led by Russel A. Kirsch at the US National Bureau of Standards. It was developed for use with a computer, and was a drum scanner.

An important concept, widely used in our paper is Computer Vision, which is a field in Artificial Intelligence, which deals with giving computers, high level understanding from digital images and videos. Computer vision deals with the automatic extraction, analysis and understanding of useful information from a single image or a series of images, also called as video.

Digital image processing was pioneered at NASA's Jet Propulsion Lab in late 1960s to convert analogue signals to digital images with computer enhancement. For a computer, an image is a 2-D signal, made of rows and columns of pixels.

In lame man's terms, in image processing, an image is processed when some transformation operations are performed on it. In technical terms, Image processing is usually related to the usage and application of mathematical functions and transformations over images. It is usually about an algorithm doing some transformations on the image such as stretching, contrasting, sharpening or smoothening on the image. While computer vision mainly focuses on image or video given as input to a computer system, and the expected output or the desired task is to analyze/infer something about the image. Computer Vision may/may not use image processing algorithms to perform this task. The goal here is to be able to gain knowledge from the image rather than enhancing the image. An enhanced image does make the process of gaining knowledge from the image easier though. So the main difference in Image Processing and Computer Vision lies in the goal. The field of Computer Vision has high potential for further development in the future by implementing machine learning algorithms to better analyze the images.

1. **Our Contribution**

The aim of this paper is to make use of Computer Vision and Python 3.7 to implement the scanning application instead of introducing heavy and expensive hardware, we create a software that does the same with a simple camera, which will easily convert hard-paper documents into scanned images, to rotate, crop (if needed) and to save these documents in the directory of the user. Also, the user gets to choose between the colored, grayscale and the usual, black and white scanned image(s).

The paper is divided into 5 sections. The first section consists of the earlier known work on this topic. Second section is about the concepts used in the making of this project. Third section is all about how the system works and the algorithms behind its working. Fourth section showcases the results generated by the system and the last section consists of the conclusion, followed by references.

## 2. Earlier Works

There has been a lot of work and research in the field of scanning and computer vision separately. The first scanner came into picture in 1975 and computer vision came to life in 1960s in the universities working on Artificial Intelligence. The key idea was to come up with a human-like visual system for the computer, so that it can see and learn for itself.

We have studied about the existing work on this project topic and tried to understand about the working and behavior of the existing systems.

The idea behind scanning application came from Boris Rewald [1], of University of Natural Resources and Life Sciences, did a comparative study of scanner based and camera based MR system or Mini Rhizotron Imaging system, Feb 2019.

Image brightness comes from Aleksey D. Poudalov, Michael lV Piltsov, Vladimir G. Mazur Angarsk[2]of state technical university,Russian Fed., in the research paper named ,"Algorithmization of the Spatial Method of Image Enhancement Based on Aligning the BrightnessHistograms",2018, giving a detailed analysis of the algorithm of the spatial method of image enhancement based on brightness histogram.

Li Xin and Shi YiLiang[3] of city college, Wuhan University Of Science and Technology did a comprehensive study, "Computer Vision Imaging using Artificial Intelligence", in 2018, of biological visual perception which included Computer Vision. They used the known laws of biological vision and evolution calculation methods to study Computer Vision. They used the perceptual model and pattern recognition, which is a core part of Computer vision, based on the evolutionary computations to discover the phenomenon and rules of biological vision.

Image segmentation and it's definition comes from M. Sridevi[4], National Institute of Technology, Tiruchirappalli, India, "Image Segmentation based on Multilevel Thresholding using Firefly Algorithm",2017, which involves identifying the distinct objects present in an image based on the properties such as intensity, color, texture etc.

The use and definition of kernels come from Krishnaprasad P, Ajeesh Ramanujan[5], "Ramanujan Sums Based Image Kernels for Computer Vision",2016.

Dr. Mostafa Abdel, Bary Ebrahim[6], in his review article, "3D Laser Scanners : History, Applications and Future", 2014, talks about the fundamentals of laser scanning, Physical principles involved , structural reconstruction and types of laser scanners. It's applications, accuracy and future of 3D laser scanners.

Slavomir Matuska, Robert Hudec, Miroslav Benco,[7] "The Comparison of CPU Time Consumption for Image Processing Algorithm in Matlab and OpenCV", 2012, concluded that OpenCv takes less CPU time for image processing than Matlab.

Adaptive threshold comes from Xiaolian Deng, Yuehua Huang, Shengqin Feng of China Three Gorges University, China and  Changyao Wang of Chinese Academy of Science, China,[8] "Adaptive Threshold Discriminating Algorithm for Remote Sensing Image Corner Detection"2010.

Pei-Ju Chiang, Nitin Khanna, Aravind K. Mikkilineni, Maria V. Ortiz Segovia, Sungjoo Suh, Jan P. Allebach,
George T.-C.Chiu, and Edward J. Delp [9], did an extensive study on the topic of "Printer and Scanner Forensics",2009, that dealt with the initial scanner and printer systems, describing architectures of various scanning and printer types introduced so far like LASER and ink-jet printers.

Peter Peer and Borut Batagelj,[10] in their paper, "Computer Vision in Contemporary Art", 2008, explored yet another striking application of Computer Vision in the field of art. The main idea behind the study being, to bring together people from Computer Vision, Graphics and Art in some kind of convergence to these topics. They presented and successfully implemented the idea of Virtual Painting Application and Virtual Skiing Application.

In this paper, we have talked about the application of Computer Vision in the field of scanning for paper-less documentation. We have successfully implemented this technique of taking image input from the webcam or from the directory of the user and editing as required to get the scanned image of the input, with up to 99% accuracy in a properly illuminated room.

3. Concepts Used
3.1. Binary Images And Threshold Images
Binary images [3] are images whose pixels have only two possible values. They are normally displayed as black and white. Numerically, the two values are often 0 for black, and 255 or 1 for white. Binary images are often produced by thresholding a grayscale or colored image (but works best for a grayscale image), in order to separate an object in the image from the background. The object with pixels 255 is referred to as the *foreground*. The rest is referred to as the *background color*.

Here we set a threshold value and all pixels greater than that value becomes white or 255 and pixels smaller than that value becomes black or 0. (In the below picture (left), we set the threshold value to be 127)

Fig.1. Grayscale Image

Fig.2. Binary Image

3.2. Structuring Element Or Kernel. This term comes from Mathematical Morphology, it is a shape used to probe or interact with a given image, with the purpose of drawing conclusions on how this shape fits or misses the shapes in the image. It is typically used in morphological operations, such as dilation, erosion, opening, and closing, as well as the hit-or-miss transform.

The structuring element is a small matrix that determines the precise details of the operator's effect on the image. The structuring element is sometimes called the kernel.

Fig.3. Example of a Kernel

The kernel slides over the image and processes the pixels that fall under it, according to the method specified by the user.

## 3.3. Perspective Transform and Contours

A perspective transform is applied to obtain a top-down, front view of the image/object placed at a reachable distance from the camera/viewer. In this we provide four points and the portion of the image lying between those four points is cropped out, and we get the top down, front view of the cropped part.

Contours can be explained simply, as a curve joining all the continuous points (along the boundary), having same pixel or intensity. The contours are a useful tool for shape inspection, object detection and object finding. There are few points to be considered for better accuracy in contour findings:

- Use binary images. So apply thresholding or canny edge detection
- In OpenCV, finding contours is like finding white object from black background. So object to be detected should be white and background should be black.

## 4. Proposed Methodology

Image Processing is used to perform operations on images, or to extract some information from them, or to get enhanced output. In this project, we use the concept of Computer Vision to scan images and give the edited scanned image as the output.

In the proposed system, we need to process our image, for which we need to take the input of the image first. There can be two ways to take the input image. (1) When the image to be scanned is already present in the directory of the user. In this case, the image can be directly read into the scanner application using OpenCv. And the other method (2) When the image to be scanned is given as input from the webcam of the user. The webcam captures a frame. The frame is then processed and then converted to desired output.

The user is given the option to select the scanned image format from threshold, colored and grayscale. The option is given in the form of switches, constructed using track-bars.

Contours of this selected image are found. (A contour is a closed loop joining all the continuous points, having same intensities).

A python list is created to store the Area(s) of all the contour(s) found and a python dictionary is created to store the calculated area(s) of the contour(s) with their corresponding contour index value(s).

Then to find the object of interest, which is to be scanned, in our input image. We find the contour of maximum area and its index.

Next, we find a rectangle of minimum area that fits the contour selected, that is, maximum area contour in the image. We use OpenCv's function to find the co-ordinates of this rectangle so formed. The points are in the order, [top left, top right, bottom left, bottom right]. The points thus found, are passed to find the perspective transform of the image, along with another set of points, which decide the alignment/positioning of the object of interest in the final image. We also create a brightened copy of this transformed image by adding '50' to each pixel of the image obtained.

The image thus found after this is sharpened and any unwanted noise in the image is removed, using various filters, like Gaussian filter, which is used to remove noise in an image. A copy of this image is converted to grayscale. Threshold of the grayscale image is calculated and bitwise not operator is applied on this image, to change the black to white and white to black again and is sharpened. Adaptive threshold is applied to another copy of the gray perspective transformed image. Bitwise and is performed between these two images to get the resulting image. The resultant image is displayed, under the title of "Scanned image".

Sometimes, while using Perspective Transform, the points returned from the boxPoint function, can be of different order, thus causing a rotation in the final image obtained. To overcome this shortcoming, we add several additional features to complete the user's experience of a scanner application.

The additional features are:

1. Pressing r/R for **right rotation**: The final image obtained is rotated right or in counter clockwise direction at 90 degree angle.

2. Pressing l/L for **left rotation**: The final image obtained is rotated left or in clockwise direction at right angle.
3. Pressing s/S for **saving** the image in the directory of the user, with the name, 'Scanned.jpg'.
4. Pressing c/C for **cropping**: Sometimes, the image obtained can't be perfectly fit into a rectangle, thus a need to crop the unwanted part, may arise. We have developed a special algorithm for cropping. Unlike the age-old method of cropping in rectangular shapes, by dragging the mouse/ finger (on a touch screen), from the top left corner to the bottom right corner of the image, we just select four points, using mouse click. The algorithm is described in brief as follows.
5. Pressing Escape or Esc to exit from the program /application.

## 4.1. FCPT Algorithm for cropping

The **'FCPT'** stands for **Four Click Perspective Transform.** The core part of this algorithm is the standard perspective transform method provided by OpenCV. In the method, we have to pass four points in a particular order like, **[Top left coordinates, Top right coordinates, Bottom left coordinates, Bottom right coordinate].** This order has to be strictly followed otherwise the output will be unpleasant. But when taking contours to separate the document/object of interest from the background, we receive four points, lying at the four corners of the document. We receives the coordinates in an unpredictable series, so in order to input the points in perspective transform function of OpenCV, there was a need of a method to identify whether the point lies in top left, top right, bottom left or bottom right corner. So we developed an algorithm which makes OpenCV's perspective transform function more user friendly by making it more predictable.

We used OpenCV's mouse call back function, which gives us the coordinate of the point where we click on the image. By using these points in our perspective transform, the cropping algorithm becomes more user friendly than the one provided by OpenCV. So in order to crop the image using perspective transform, the user only has to click the four corners of the document in any order.

The above coordinates are then received by a python list, where the point determination method identifies whether the point lies on the top left, top right, bottom left or bottom right corner of the document. After determining the position, the coordinates are passed into the perspective transform function of OpenCV and we get the result, that is, the "cropped image" is displayed.

The flowchart below depicts the working/functionality of the algorithm stated and explained in the fourth section. The algorithm is stated and can be understood easily using this precise flowchart. The major steps involved are:
1. Loading an image in colored and grayscale.
2. Calculate Perspective Transform and find Contours.
3. Performing the required operations on the image to enhance the features of the image.
4. Performing Adaptive Thresholding and Bitwise operations to further enhance the image(s).

The user is given the option to choose between the colored/grayscale/black and white (threshed) scanned image. He/She can also perform operations like rotation and cropping on the scanned output image, which further adds to the experience of the user.

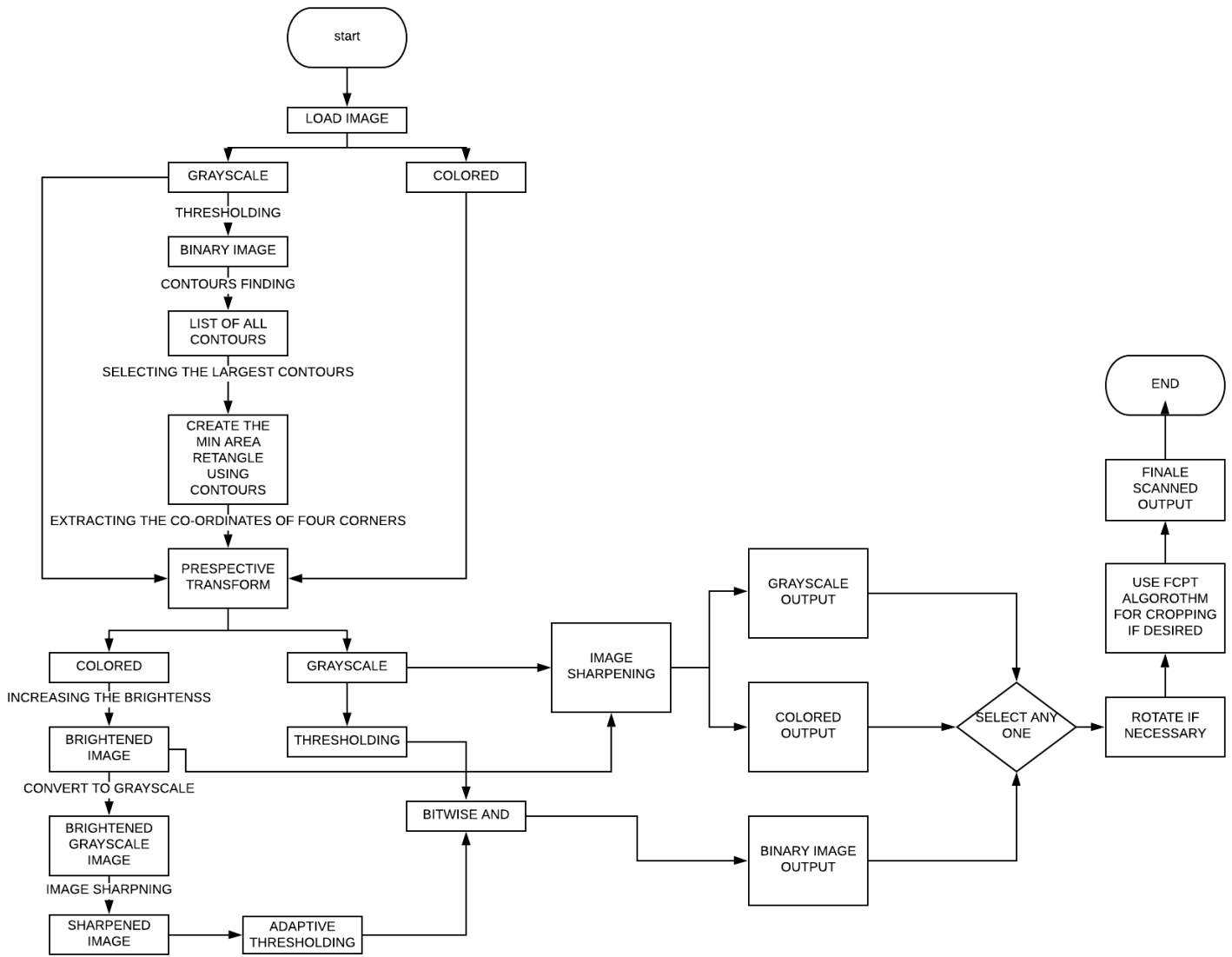

**Fig.4**. Flowchart to represent the algorithm used for the application/program based on OpenCv, an open source library.

## 5. Results and Discussions

The proposed system was tested on a number of documents/images to find their scanned images as outputs. They are shown as follows:

(1) We give a simple but dim-lit notice as our input image. The original image is displayed in fig. 5. We get the scanned copy of our input image, which is labeled as 'Scanned'. The user gets to choose between the thresh, colored and grayscale image. Thresh image is shown in fig.6. But as the 'Scanned' copy came out to be rotated, we need to rotate the image in the anti-clock wise direction, for which we press l/L. The result is another image pop-up, labeled as 'Rotated Image', shown in fig.7.We can also crop this rotated image, for removing the extra part, for which we press c/C. fig. 8. Shows the cropped image.

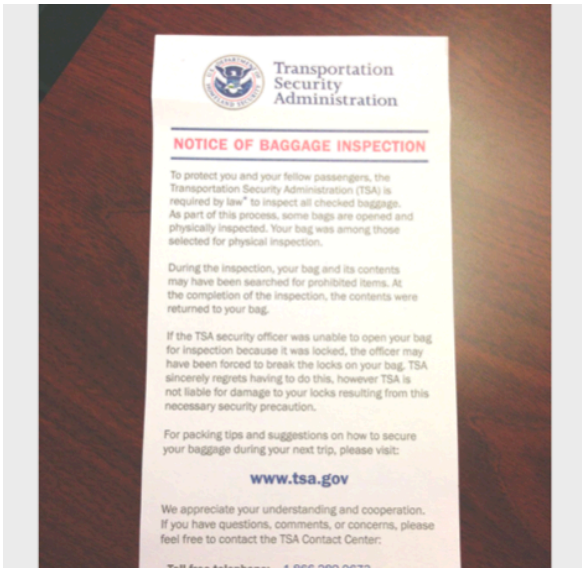

**Fig.5.** original image of the notice

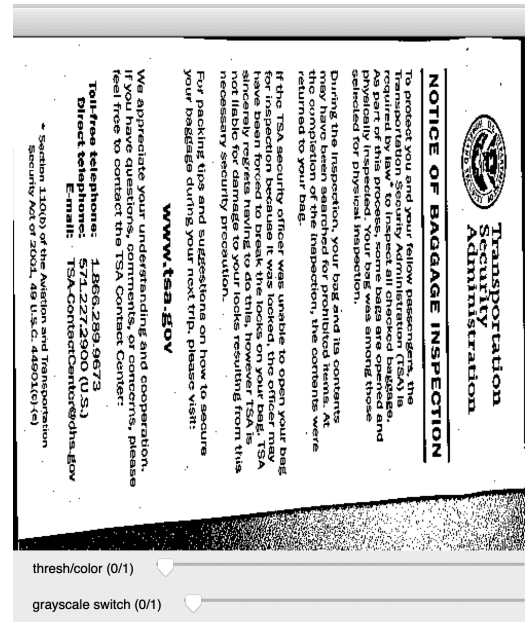

**Fig.6.** Scanned image

This is the final image which has been scanned, rotated and then cropped is shown in fig. 8. The enhancement of readability can be seen.

This image can be saved by pressing 's' or 'S', and it will be saved in the directory of the user.
The effectiveness of the cropping algorithm devised can also be seen substantially. The unwanted blackened part of the image, in the rotated image is cropped up using only four clicks!

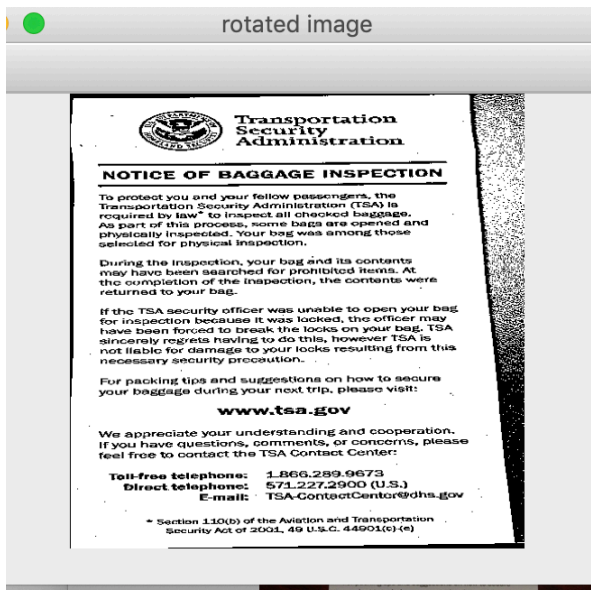

**Fig.7.** Rotated image

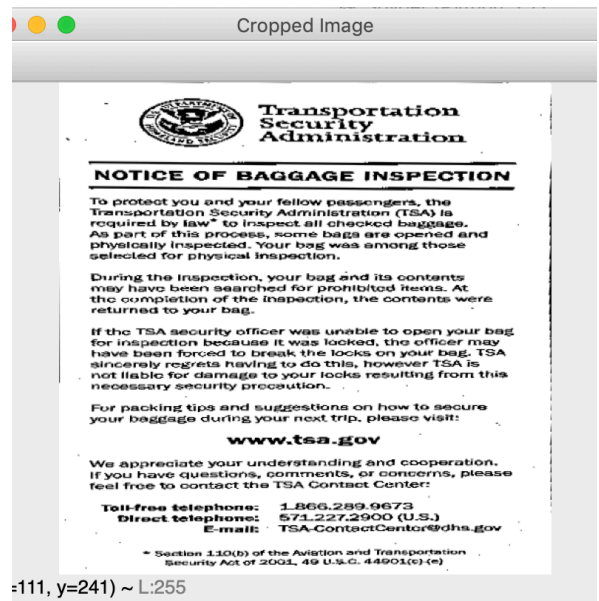

**Fig.8.** Cropped image

(2) Next we feed an identity card called as the Aadhar card as our input image. The card is lying on the table in a titled manner, as shown in fig.9. Again the user selects between the thresh, colored and grayscale images, labeled in a window called 'Scanned Image' in fig.11.

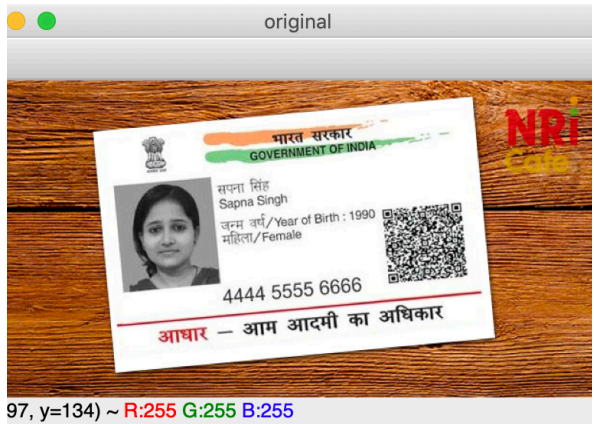

**Fig.9.** Original image

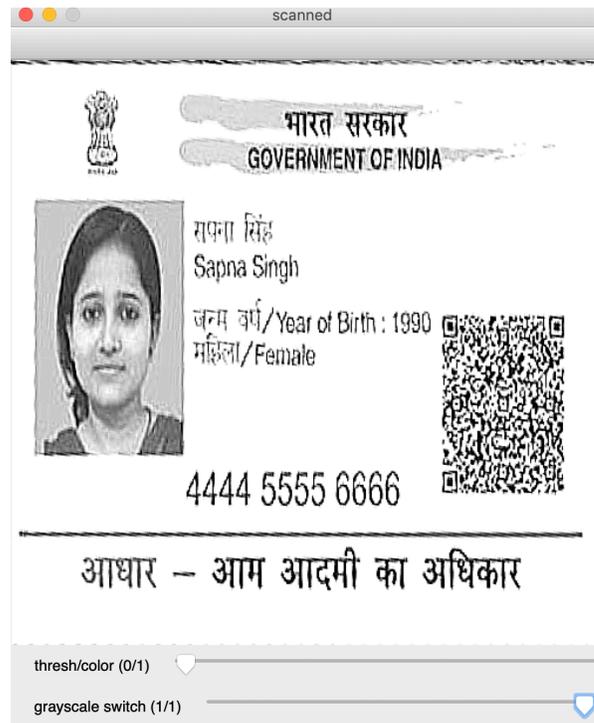

**Fig.10.** Grayscale scanned image

The given identity card is first converted to the scanned output image, for the user to explore his/her options. The scanned image can be converted to grayscale, as shown in fig. 10 or it can be converted to a colored image, as shown in fig. 12.

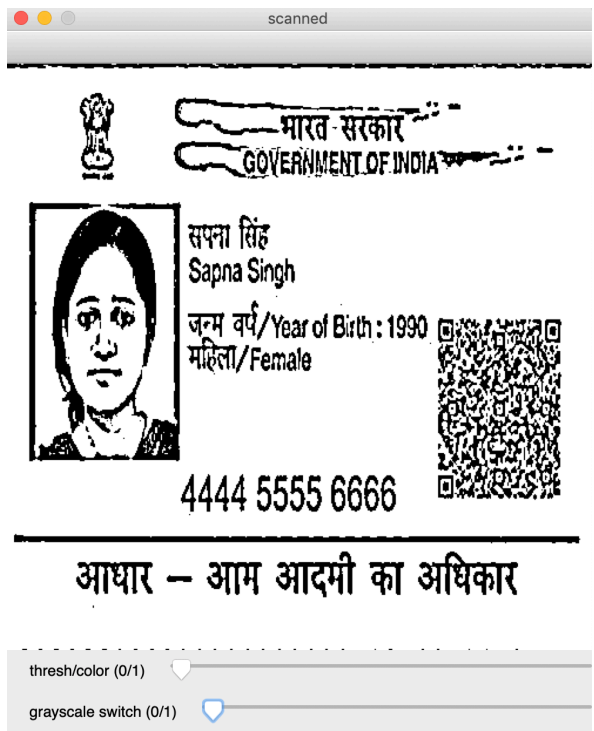

**Fig.11.** Scanned thresh image

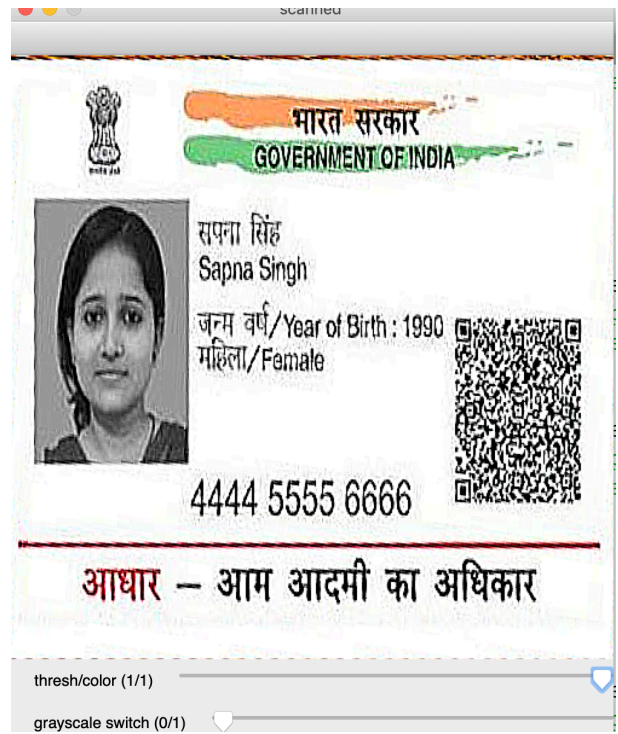

**Fig.12.** Colored scanned image

(3) Here we give the input image using the webcam of the system. The input original image is shown in fig.13. The scanned output is shown in fig.14, labeled in a window 'Scanned Image'. The input image is beautifully cropped up and zoomed. Though the limitations of the webcam affect the quality of the scanned image, this can be avoided by using a better camera.

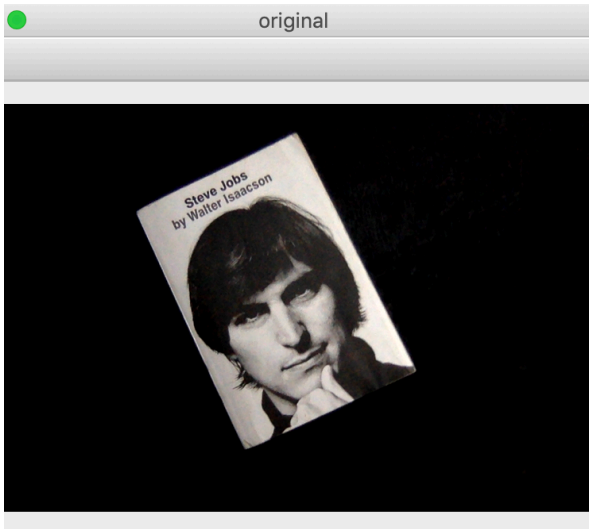

**Fig.13.** Original image from the webcam

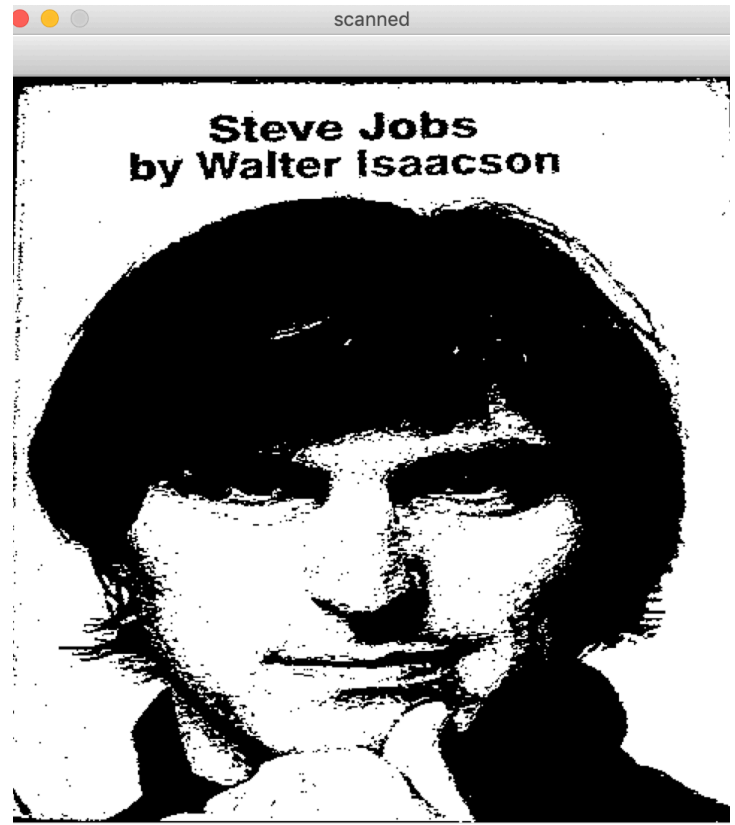

**Fig.14.** Scanned image

### 6. Conclusion

We proposed a computer vision based scanning method, made robust by thresholding, track-bars as switches for thresh/color/grayscale image options, perspective transform, brightening of image, rotation functionalities, bitwise operators, Adaptive thresholding and a special cropping algorithm. We tested our proposed idea on several sample documents successfully. We can make it better by adding several other filters for color enhancement and better readability or introduce OCR (Optical Character Recognition) using Machine Learning to add-on to the user experience.